\documentclass[conference]{IEEEtran}
\IEEEoverridecommandlockouts
\usepackage{cite}
\usepackage{amsmath,amssymb,amsfonts}
\usepackage{graphicx}
\usepackage{textcomp}
\usepackage{xcolor}
\usepackage{epsfig}
\usepackage{amsmath}
\usepackage{amssymb}
\usepackage{color}
\usepackage{url}
\usepackage{multirow}
\newlength{\figurewidth}
\newlength{\smallfigurewidth}
\newcommand{\eatme}[1]{ }
\usepackage{paralist}
\usepackage{adjustbox}
\usepackage{wrapfig}

\usepackage{cite}
\usepackage{amsmath,amssymb,amsfonts}
\usepackage{algorithm}
\usepackage{graphicx}
\usepackage{textcomp}
\usepackage{xcolor}
\usepackage{subcaption}
\usepackage{siunitx}

\usepackage[T1]{fontenc}
\usepackage{graphicx}
\usepackage{booktabs}
\usepackage{longtable}
\usepackage{pbox}
\usepackage{mathtools}
\usepackage{caption}
\usepackage{subcaption}
\usepackage{algpseudocode}
\usepackage{algorithm}
\usepackage{siunitx}
\usepackage[inline]{enumitem}
\usepackage{siunitx}
\usepackage{diagbox}
\usepackage[]{mdframed}

\usepackage{comment}
\usepackage[normalem]{ulem}

\DeclarePairedDelimiter\abs{\lvert}{\rvert}

\usepackage{xparse}
\NewDocumentCommand\N{sm}{\mathcal{N}\IfBooleanT#1{^{\ast}}_{#2}}

\makeatletter
\NewDocumentCommand{\@Coefficients}{m}{\text{\ttfamily\upshape #1}}
\newcommand\uMultilevelCoefficients{\@Coefficients{u\char`_mc}}
\newcommand\newMultilevelCoefficients{\@Coefficients{\~{u}\char`_mc}}
\makeatother

\NewDocumentCommand\at{m}{\text{\upshape\ttfamily\lbrack}#1\text{\upshape\ttfamily\rbrack}}

\begin{document}

\title
{\large
\textbf{Machine Learning Techniques for Data Reduction of Climate Applications \thanks{This work was partially supported by DOE RAPIDS2 DE-SC0021320 and DOE DE-SC0022265.}}
} 

\author{%
Xiao Li$^{\ast}$,   Qian Gong$^{\dag}$, Jaemoon Lee$^{\ast}$, Scott Klasky$^{\dag}$, Anand Rangarajan$^{\ast}$ \\ Sanjay Ranka$^{\ast}$\\[0.5em]
{\small\begin{minipage}{\linewidth}\begin{center}
\begin{tabular}{ccc}
$^{\ast}$University of Florida & \hspace*{0.5in} & $^{\dag}$Oak Ridge National Laboratory \\
\end{tabular}
\end{center}\end{minipage}}
}

\maketitle
\thispagestyle{empty}

\begin{abstract}
Scientists conduct large-scale simulations to compute derived quantities-of-interest (QoI) from primary data. 
Often, QoI are linked to specific features, regions, or time intervals, such that data can be adaptively reduced without compromising the integrity of QoI. For many spatiotemporal applications, these QoI are binary in nature and represent presence or absence of a physical phenomenon. 

We present a pipelined compression approach that first uses neural-network-based techniques to derive regions where QoI are highly likely to be present. Then, we employ a Guaranteed Autoencoder (GAE) to compress data with differential error bounds. GAE uses QoI information to apply low-error compression to only these regions. This results in overall high compression ratios while still achieving downstream goals of simulation or data collections. Experimental results are presented for climate data generated from the E3SM Simulation model for downstream quantities such as tropical cyclone and atmospheric river detection and tracking. These results show that our approach is superior to comparable methods in the literature.
\end{abstract}

\section{INTRODUCTION}

Scientific applications are producing data at an unprecedented rate, with both volume and velocity expanding rapidly, outpacing advancements in storage, computation, and network capacities~\cite{zou2014improving, wen2018compression}. This growth is also seen in next-generation experimental and observational facilities, making data reduction  an essential functionality because it's often impractical to retain data at the original scale. Data reduction necessitates the development of algorithms that can reconstruct data with rigorous quality assurance so that application scientists can trust the reduced data for use in their analyses. 

Because the errors on derived quantities or quantities of interest (QoI) may deviate from these on lossy reduced primary data (PD), it is important that reduction techniques can maintain the accuracy of QoI and provide high compression ratios. Long-term and large-scale scientific simulations often model dynamic processes over an extended period and write out field data at time steps. In this paper,  we target the spatiotemporal data analysis with climate simulation, as the presence or absence of a physical phenomenon (i.e., QoI) can be represented by binary masks. Throughout this paper, we refer to locations where climate features are presented as \textit{critical regions} or \textit{regions-of-interest} (ROIs). Our goal in this paper is to demonstrate a hybrid, feature-driven compression approach. The proposed pipeline consists of three stages (Figure \ref{fig:gae}):
\begin{compactitem}
\item
ROI Detection: We predict regions where these QoI are likely to be present. We employ historical data sets and a customized UNet model. The output is binary masks that indicate the presence of climate phenomena of interest.
In this work, we propose to hybridize neural network (NN) prediction with differential compression. Our approach relies on a UNet-based neural network that can jointly predict ROIs for different climate events, such as AR and TC.  Our neural network model generates probability maps for climate events where climate ROIs may be present. The allocation of ROIs can be fine-tuned by adjusting the acceptance threshold. This approach offers a convenient way to balance between false positive rates and compression ratios.
\item 

Guaranteed Autoencoder: The original data are divided into smaller spatiotemporal blocks. a 3D convolutional autoencoder is utilized to capture the spatiotemporal correlations within each block. To guarantee the error bound of the reconstructed data, Principal Component Analysis (PCA) is applied to the residual between the original and reconstructed data. This yields a basis matrix, which is then used to project the residual of each instance. The resulting coefficients are retained to enable accurate recovery of the residual. The number of coefficients saved is incrementally increased until the error bound is satisfied. Additionally, quantization and entropy coding techniques are applied to both the latent data from the GAE and the PCA coefficients. This further improves the compression ratio of the overall process.

\item
Differential compression: We use lower error bounds on predicted critical regions and higher error bounds on the rest of space to ensure minimal false negatives on QoI and large compression ratios globally.
Currently, most state-of-the-art lossy compressors \cite{di2016fast, gong2022region, lindstrom2014fixed} are mostly focused on controlling global error such as $L^\infty$ and $L^2$. Without local error control, the application of differential compression requires the partitioning of ROIs and the rest of the data and compression of each part separately. In this paper, we employ a multilevel-decomposition-based compression method discussed in \cite{gong2022region} because it supports pixel-wise changed error bounds. The differentially compressed data can be reconstructed using the same algorithm and code as the uniformly compressed data. 
\end{compactitem}
We validate our hybrid method using the simulation data generated by Energy Exascale Earth System Model (E3SM)~\cite{golaz2019doe} and assess the impact of compression errors on two climate QoI: tropical cyclone (TC)~\cite{balaguru2020characterizing} and atmospheric river (AR) \cite{kim2022atmospheric} tracking. 
In comparison to previous research \cite{gong2022region}, our method achieves a substantially higher compression ratio while also attaining a lower false negative (FN) ratio for both TC and AR detection, along with significantly improved quality of information preservation.

\eatme{
Our work targets TC and AR as they are impactful weather phenomena that can lead to substantial loss of life and economic damage \cite{carleton2016social, clarke2022extreme}. Their detection requires detection of candidates at individual snapshots, stitched across time steps. 

The previous work in \cite{gong2022region} uses the residual of multilinear interpolation -- an intermediate outcome of the multilevel compression -- and an adaptive-mesh-refinement method to identify candidate TC and AR regions. Because the residual is a general-purpose, analysis-agnostic metric measuring local data variation, experiments show that the previous work overestimates the share of ROIs, leading to suboptimal compression ratios.


\item It leverages traditional techniques for guaranteeing error on primary data from scientific applications. Examples of such techniques include MGARD\cite{ainsworth2019multilevel}, FPZIP\cite{lindstrom2006fast}, and ZFP\cite{lindstrom2014fixed}. 
\end{compactenum}
}


The rest of the paper is organized as follows. Section ~\ref{methodology} describes our compression pipeline. Experimental results using our method are presented in Section~\ref{expt}  for different levels of compression and accuracy. Section~\ref{relwork} presents related work. Conclusions are provided in Section~\ref{conclusion}.



\section{Methodology}\label{methodology}

\begin{figure}
  \centering
    \includegraphics[width=1\linewidth]{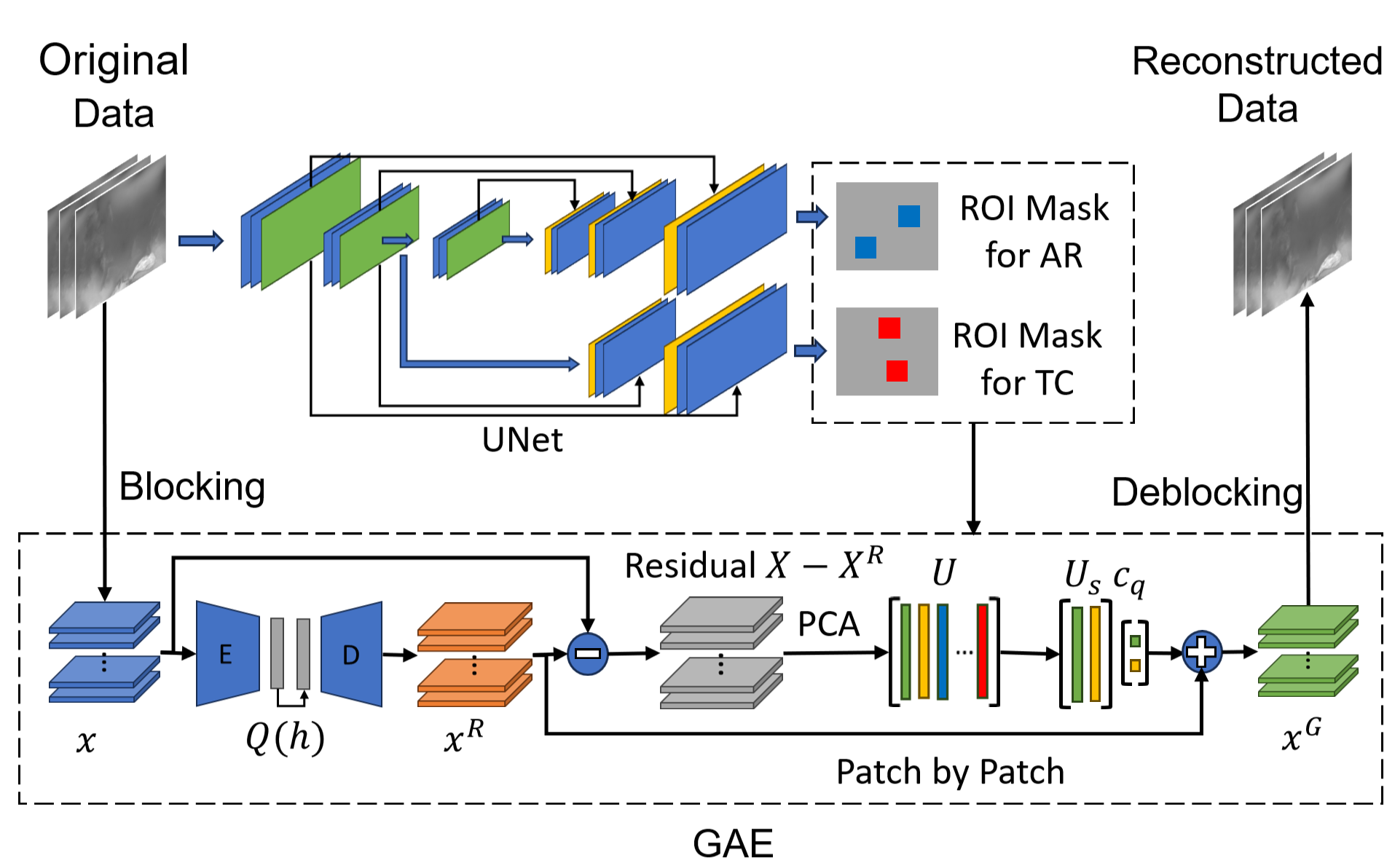}
    \caption{Method Overview: $x$ denotes AE the input block, while $x^R$ denotes AE the output block. $x^G$ denotes the error-bounded output block. $Q(h)$ represents latent space quantization. $U$ denotes PCA basis matrix. $U_s$ and $C_q$ are selected basis vectors and corresponding quantized coefficients, respectively.}
    \label{fig:gae}
    \vspace*{-0.5cm}
\end{figure}

The overview of our compression approach is given in Figure \ref{fig:gae}. Our method includes two parts: ROI detection and GAE. A UNet is employed for the detection of ROIs. We perform differential compression with GAE by applying lower error bound for ROIs and higher error bound for other regions. In the following, we briefly describe the ROI detection model and the GAE.

    



\paragraph*{ROI Detection}
We used a UNet \cite{ronneberger2015u} architecture with a U-shaped interconnection topology, which includes a contracting pathway to capture contextual information and an expanding pathway for accurate localization. Our network accepts input data with dimensions of $C \times H \times W$, where $C$ denotes the number of climate variables, while $H$ and $W$ correspond to the height and width of the input images, respectively. 
To meet the requirement of jointly predicting the locations of TC and AR, we have made the following revisions to the original UNet architecture and extended the decoder into two branches (Figure \ref{fig:gae}) so that the two decoders share the same encoder. We use a shallow branch for TC detection, while using a deep branch for AR detection. The reason for this design choice is that we have found that using a deep neural network can yield better results for AR, while a shallow network is sufficient for TC detection. The output of each decoder is passed through a sigmoid function to constrain the output within the range $[0, 1]$. The network's output takes the form of a confidence map with dimensions $H \times W$, effectively represented as an image comprising probability values. These probabilities can then be suitably thresholded to derive ROIs.


Given that TCs are relatively rare and infrequent phenomena on Earth, it can be challenging for the model to learn robust features under the supervision of very sparse coordinates. Therefore, we convert the TC's location coordinates into a Gaussian heatmap using the provided equation:

\begin{equation}
    H_i(u, v) = A \cdot \exp\left(-\frac{(u - u_i)^2 + (v - v_i)^2}{2\sigma^2} \right),
\end{equation}
where $(u_i,v_i)$ denotes the location of the TC in the image coordinate system, $\sigma$ represents the standard deviation controlling the spread of the TC representation, and A is a scaling factor that controls the overall peak intensity of the heatmap.

Because there could be more than one TC in a single data frame, we generate a heatmap for each cyclone within one frame and then combine them into a single heatmap using the following equation:

\begin{equation}
    H_{\text{combined}}(x, y) = \max_{i=1}^{N} H_i(x, y),
\end{equation}
where $N$ is the number of TCs in one frame. The combined heatmap serves as the ground-truth label for model training.


The locations of ARs are represented using binary masks, where each pixel indicates the presence or absence of an AR at the corresponding location.

\paragraph*{GAE}
An autoencoder (AE) is a neural network designed to learn efficient representations of data by compressing the input into a lower-dimensional latent space and then reconstructing the original input from this compressed representation. It consists of two main parts: an encoder and a decoder. The encoder compresses the input data from high-dimensional to lower-dimensional latent space representation, while the decoder reconstructs the original input from this compressed representation. \cite{jaemoon}

AEs can be trained by comparing the uncompressed and reconstructed data. The initial step towards achieving an effective codec involves minimizing the reconstruction error adequately. In this study, we employ the standard mean-squared error (MSE) loss function to measure compression-incurred errors in PD. 


After training, the decoder and the latent representations  $\boldsymbol{H} = \left\{\boldsymbol{h}_{i}\right\}_{i=1}^{N}$ need to be stored. We store the decoder without compression, given its small size.  However, storing floating-point latent vectors is not an efficient approach. To overcome this challenge, we employ a compression technique that involves float-point quantization followed by entropy encoding to compress the latent space data. Our approach to enhancing compression efficiency employs a two-step strategy. Firstly, we uniformly quantize these coefficients into discrete bins, each with a bin size of $d$. This discretization process effectively represents all values within each bin by its central value, transforming the originally continuous data into a discrete form. Subsequently, we utilize Huffman coding to compress these quantized coefficients. Huffman coding assigns shorter codes to frequently occurring quantized coefficients, optimizing the representation of the data and achieving higher compression efficiency. This combined technique significantly reduces the data's size while retaining crucial information.

\begin{figure}
  \centering
    \includegraphics[width=0.9\linewidth]{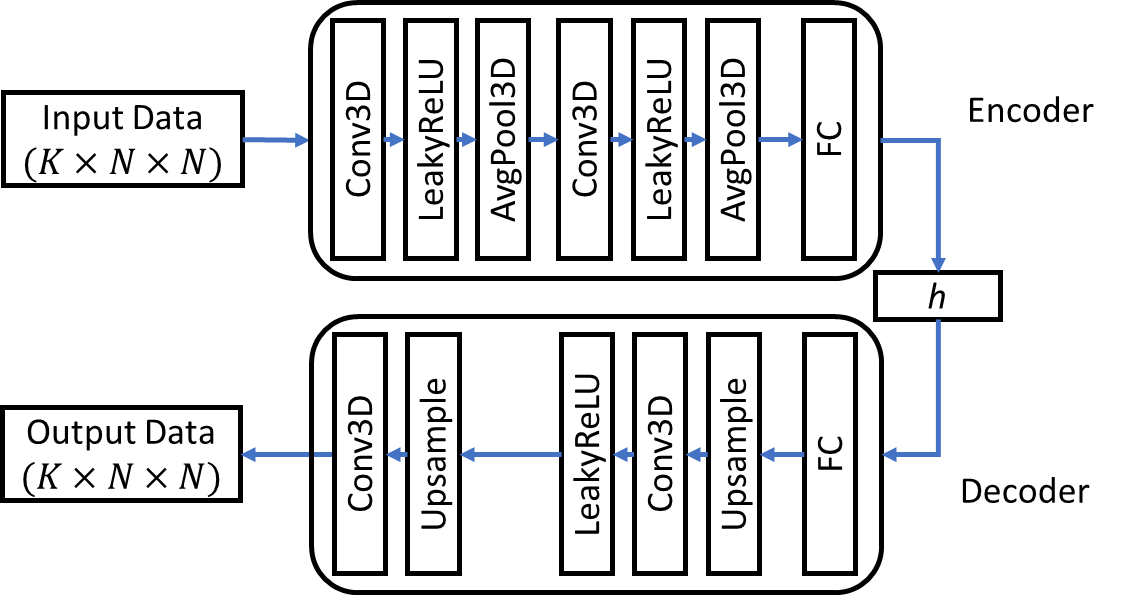}
    \caption{The structure of the autoencoder: Conv3D denotes the 3D convolution layer, FC denotes the fully connected layer, and $h$ denotes the latent space. Leaky ReLU is adopted as the activation function. We use a 3D average pooling layer to reduce the size of features and trilinear upsampling for interpolation.}
    \label{fig:ae}
     \vspace*{-0.5cm}
\end{figure}

\paragraph*{Bounding the Reconstruction Error}

We aim to limit reconstruction errors for all instances in an AE. Although any appropriate error bound can be applied within our framework, our emphasis lies in constraining the $\ell_{2}$ norm of each instance residual, denoted as $\left\|\boldsymbol{x}-\boldsymbol{x}^{R}\right \|_{2}$. To optimize compression ratios, we only bound the AE reconstruction error for instances whose residual $\ell{2}$ norm exceeds the specified threshold $\tau$.
After obtaining the reconstructed data from the autoencoder, we apply Principal Component Analysis (PCA) to the residual of the entire dataset to extract the principal components or basis matrix, denoted as 
${U}$. These basis vectors are sorted in descending order according to their corresponding eigenvalues. These principal components capture the directions of maximum variance in the residual data. Although we compress the data block by block, we treat each patch of data as a single instance and compute the basis matrix at the patch level. To guarantee the error bound for each patch of data, we project the residual of each patch of data onto the space spanned by $U$ and select the leading coefficients such that the $\ell_2$ norm of the corrected residual falls below the specified threshold $\tau$. These coefficients, representing the residual, are derived from the equation: 
\begin{equation}
\boldsymbol{c}=U^{T}\left(\boldsymbol{x}-\boldsymbol{x}^{R}\right).
\end{equation}
It's important to note that the complete recovery of the residual $\boldsymbol{x}-\boldsymbol{x}^{R}$ can be achieved by computing $\boldsymbol{Uc}$, yielding the coefficient vector $\boldsymbol{c}\equiv \left[c_{1},\ldots,c_{D}\right]$.  Given that the error bound criterion is based on $\ell_{2}$, we compute $\{c_{k}^{2}\}_{k=1}^{D}$ and sort the positive values. The coefficients are therefore sorted in the order of largest contribution to the error. The top $M$ coefficients and corresponding basis vectors are selected to satisfy the target error bound $\tau$. To minimize the storage cost of these coefficients, we compress the selected coefficients $\boldsymbol{c}_{s}$ using a method similar to that employed for compressing AE latent coefficients. These coefficients are quantized before being used for reconstructing the residual. The corrected reconstruction $x^{G}$ is
\begin{equation}
\boldsymbol{x}^{G}=\boldsymbol{x}^{R}+U_{s}\boldsymbol{c}_{q},
\end{equation}
where $\boldsymbol{c}_{q}$ is the set of selected coefficients $\boldsymbol{c}_{s}$ after quantization and $U_{s}$ is the set of selected basis vectors. We increase the number of coefficients until we achieve $\left\|\boldsymbol{x}-\boldsymbol{x}^{G}\right\|_{2}\leq \tau$. Therefore, in order to guarantee the error bound, we need to store the basis matrix, the selected coefficients $\boldsymbol{c}_{q}$ for each patch data and their basis vector indices.

To explore the spatiotemporal correlation within scientific data, we integrate 3D convolution into our AE architecture that incorporates both spatial and temporal correlations.
By harnessing the capabilities of 3D convolutional operations, our autoencoder excels in capturing intricate spatial patterns and temporal dynamics simultaneously. We partition the original data into non-overlapping $ N \times N$ patches at each data frame. Then, we group $K$ consecutive patches from the same location across time into a single block of data as an input to the AE. Each block of data is processed independently by the AE. The structure of the AE is shown in Figure \ref{fig:ae}.

\begin{figure}
  \centering
    \includegraphics[width=0.9\linewidth]{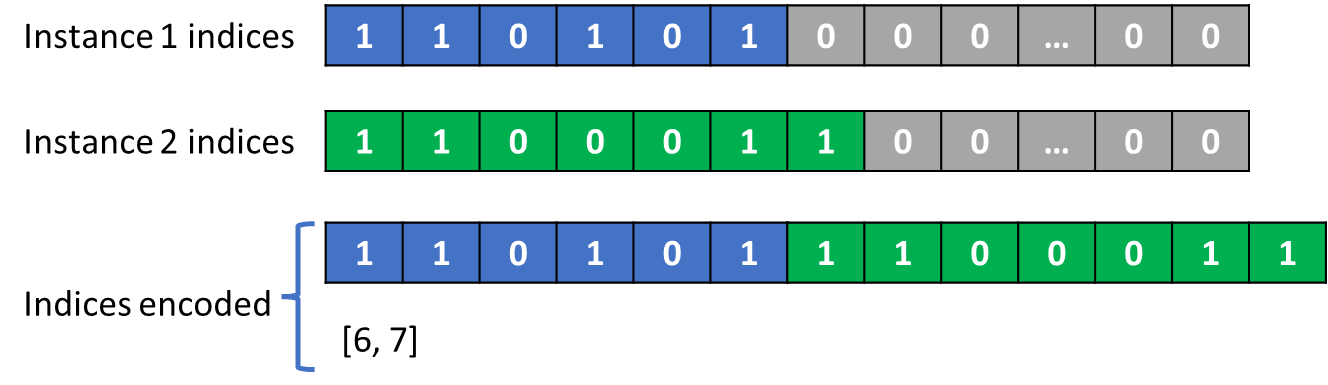}
    \caption{Indices encoding}
    \label{fig:indices}
     \vspace*{-0.5cm}
\end{figure}

To store the basis vector indices for each patch of data effectively, we start by encoding the indices using a binary sequence. Remarkably, we notice that the initial basis vectors (associated with larger eigenvalues) are selected more frequently. As a result, these binary sequences frequently end with a sequence of zeros. We decide to store only the shortest prefix subsequence that contains all ones. Furthermore, we introduce an additional value to indicate the length of this subsequence. An illustrative example is provided in Figure \ref{fig:indices}.

To guarantee QoI while maintaining a high compression ratio, we categorize all patches of data into three classes, each with a distinct error bound: ROI patches, buffer zone patches, and non-ROI patches. These are based on thresholding the output of the previously described UNet model that produces binary masks indicating the presence or absence of TC/AR. To determine if a patch belongs to the ROI category, we evaluate the proportion of pixels predicted as ROI within each patch data. If this percentage surpasses a specific threshold, the patch is designated as an ROI patch. Furthermore, within the non-ROI patches, we identify neighboring ROI patches as a Buffer Zone patch. This procedure generates a set of error bounds for each patch. Details of the overall approach are described in Algorithm~\ref{alg:GAE}.

\begin{algorithm}
\caption{The GAE Algorithm}
\label{alg:GAE}
\begin{flushleft}
\textbf{Input:} Input data $\boldsymbol{X} = \left\{\boldsymbol{x}_{i}\right\}_{i=1}^{N}$, reconstructed data $\boldsymbol{X}^R = \left\{\boldsymbol{x}_{i}^{R}\right\}_{i=1}^{N}$, target error bounds $\left\{\tau_{i}\right\}_{i=1}^{N}$.

\textbf{Output:} Corrected reconstruction $\boldsymbol{X}^G = \left\{\boldsymbol{x}^{G}_{i}\right\}_{i=1}^{N}$, coefficients $\mathbf{C} = \left\{\boldsymbol{c}_{i}\right\}_{i=1}^{N}$, indices $\mathbf{I} = \left\{I_{i}\right\}_{i=1}^{N}$ where $I_i$ is an index set, basis matrix $U$.
\end{flushleft}
\begin{algorithmic}[1]
\State Run PCA on the residual $\boldsymbol{X}-\boldsymbol{X}^R$, obtaining basis matrix $U$
\For{$i=1$ \textbf{to} $N$}  
\State $\boldsymbol{x}\gets \boldsymbol{x}_{i}$, $\boldsymbol{x}^{R}\gets \boldsymbol{x}_{i}^{R}, \tau \gets \tau_i$.
\State Compute $\ell_{2}$ norm $\delta=\left\|\boldsymbol{x}-\boldsymbol{x}^{R}\right\|_{2}$.
\If{$\delta > \tau$}
\State Project residual $\boldsymbol{c}=U^{T}(\boldsymbol{x}-\boldsymbol{x}^{R})$ and sort $c_{k}^{2},~\forall k$.
\State $M\gets 1$
\While{$\delta>\tau$}
\State $\boldsymbol{c}_{s},U_{s}\gets$ Top $M$ coefficients in $\boldsymbol{c}$ and corresponding basis vectors in $U$.
\State $\boldsymbol{c}_{q}$ $\gets$ Quantize($\boldsymbol{c}_{s}$)
\State $\boldsymbol{x}^{G}\gets \boldsymbol{x}^{R}+U_{s}\boldsymbol{c}_{q}$.
\State $\delta \gets \left\|\boldsymbol{x}-\boldsymbol{x}^G\right\|_{2}$
\State $M\gets M+1$
\EndWhile
\State $\boldsymbol{c}_{i}\gets \boldsymbol{c}_{q}$
\State $I_{i}\gets$ Index set for $\boldsymbol{c}_{q}$
\State $\boldsymbol{x}^{G}_{i}\gets \boldsymbol{x}^{G}$
\EndIf
\EndFor
\end{algorithmic}
\end{algorithm}



\paragraph*{MGARD}
We compare our method with a state-of-the-art lossy compressor -- MGARD\cite{ainsworth2019multilevel}, as it can take in customized ROI masks then conduct regionally-varied, error-bounded compression. MGARD utilizes a multilevel decomposition-based compression algorithm and mathematically ensures the errors induced by lossy compression to stay below user-prescribed error tolerance. MGARD treats input data $u$ in $d$ dimensions as the values taken by a continuous function on a grid $N_L$. It decomposes $u$ into a set of multilevel coefficients $\uMultilevelCoefficients$, residing on a hierarchy of subgrids $\N{L - 1}, \dotsc, \N{l}, \dotsc, \N{0}$ downsampled from $N_L$ through $L^2$ projection and multilinear interpolation operations. MGARD then leverages linear-scaling quantization, Huffman encoding \cite{moffat2019huffman}, and GZIP/ZSTD compression \cite{abdelfattah2014gzip,collet2016smaller} to convert $\uMultilevelCoefficients$ into a reduced representation subject to user-prescribed error bounds.

\eatme{
Due to the multilevel decomposition algorithm employed by MGARD, point-wise compression errors will spread to the entire space after recomposition. The work in \cite{gong2022region} proves that for any uniform grid space, the compression error induced by quantizing a multilevel coefficient $\uMultilevelCoefficients$ at the node $x$ at level $l$ will decay at an exponential rate to the grid spacing at the next coarser level:
\begin{equation}
\abs{(\texttt{u} - \tilde{\texttt{u}})(y)}
\leq C_d \sum_{l = 0}^{L} \sum_{x \in \N*{l}} \abs[\big]{\Delta\uMultilevelCoefficients\at{x}} 
(2+\sqrt{3})^{-d_{l-1}(x, y)}.
\label{eq:point-wise-error}
\end{equation}
Here, $\Delta\uMultilevelCoefficients\at{x}$ is the quantization error occurred at $\uMultilevelCoefficients\at{x}$, $C_d$ is a constant related to the data dimension, and \(d_{l-1}(x, y)\) represents the number of grid spacing between $x$ and $y$ at the next coarser level. The above equation suggests that the regional compression errors are chiefly attributed to the errors in the region itself and a thin surround areas. Accordingly, the work in \cite{gong2022region} proposes a buffer-zone based approach to control MGARD compression errors in the ROI. For comparison, we also feed the ROI masks generated by the UNet model into the MGARD pipeline, and compress data in both ROIs and derived buffer zones with lower error bounds.

}

\section{Experimental Results}\label{expt}
T his section presents the experimental results of compressing the E3SM through our pipeline. We  provide a concise description of the data to be compressed and evaluation metrics, followed by a comparison of accuracy and compression ratio achieved by our method as compared to previous work.

\paragraph*{Dataset}
The testing data come from a HR configuration atmosphere  E3SM simulation spanning one year. In this simulation, E3SM utilizes a grid resolution of 25 km, resulting in a total of 350,000 float32 data points per variable per time-slice. The grid spacing used in the data simulation is $0.25^{\circ}$. The TC and AR detection are treated as blackboxes in our approach. We used TempestExtremes version 2.1 \cite{ullrich2021tempestextremes}, a widely recognized software package designed for feature tracking and scientific analysis of global earth-system data. The detected TCs are nodal points in latitude and longitude coordinates, and the atmospheric rivers (AR) are areas formed by connected grid points. Eligible candidate across consecutive timesteps can be stitched to form TC and AR trajectories.   
The outcomes of TempestExtremes for the AR and TC locations at individual snapshot (i.e., timestep output) are used as binary masks and serve as the ground-truth labels for both model training and testing purposes.


\paragraph*{UNet Configure} To achieve better compression ratios, for each snapshot, we transform variables from their original 1{\scshape d} grids to three 2{\scshape d} snapshots correlated in longitude and latitude on the cubed sphere. We split the dataset into two subsets: a training set and a testing set, which include 34,560 images and 21,600 images, respectively. We use the following  TC tracking variables as inputs:  pressure at sea level (PSL), temperature at 500 hPa and 200 hPa (T500, T200), meridional and zonal wind speed at surface (VBOT, UBOT), and 1 AR tracking variable --integrated vapor transport ($\text{IVT} = \sqrt{\text{TVQ}^2+\text{TUQ}^2}$, where TVQ and TUQ denote meridional and zonal water flux, respectively)  into the developed model and simultaneously make predictions on the above two QoI. 
\paragraph*{AE Configure} To train the autoencoder, we partition the dataset into uniform blocks. Experimentally, each block is defined by a size of $8 \times 16\times 16$, where the spatial dimensions are $16\times16$, and data from 8 consecutive frames at the same location are combined into a single block. Using the mean squared error (MSE) loss function to train the network, our method is assessed across all the climate variables: PSL, T200, T500, VBOT, UBOT, TVQ, and TUQ. The first five variables serve for TC detection, while the latter two are utilized for AR detection. Separate autoencoders are trained for each variable.

\paragraph*{Evaluation Metrics}

We evaluate our method from three aspects: ROI detection, QoI preservation, and primary data error. The overall compression ratio that can be achieved is directly proportion to the percentage of total area or volume covered by ROIs as well as the false negatives on the QoI.
 For a given QoI, the FN percentage reflects the proportion of incorrectly classifying genuine ROIs as negative. The ROI ratio indicates the portion of the region designated as ROIs based on prediction results. This assessment helps us gauge the model's effectiveness in accurately detecting and encompassing all ROIs. 
 The resulting reconstructed climate data are then inputted into TempestExtremes to generate AR and TC predictions. Simultaneously, we also run TempestExtremes on the uncompressed climate data. This is done by a  comparison between the detection results obtained using reconstructed data and original uncompressed data.
 
 For TC evaluation, we compare the locations of cyclones obtained from the compressed and uncompressed data. we use TC error rate to measure the performance of our method on TC preservation. The TC error rate is defined as:

\begin{equation}
\text{TC Error Rate} = \frac{N_{error}}{N_{total}} * 100 \%
\end{equation}
where $N_{total}$ represents the total count of ground truth TCs, and $N_{error}$ denotes the count of TCs from the compressed data that deviate from their ground truth locations. 


For AR evaluation, we adopt the Jaccard index or intersection-over-union ($\text{IoU}$) to evaluate the performance of our method. $\text{IoU}$ can be expressed as
\begin{equation}
    \text{IoU} = \frac{\text{TP}}{\text{TP}+\text{FP}+\text{FN}},
\end{equation}
where $\text{TP}$, $\text{FP}$, $\text{FN}$ denote the number of true positive, false positive, and false negative predictions for AR.

For primary data, we utilize the relative \( l_2 \) norm as the evaluation metric, defined as:

\[
\text{Relative } l_2 = \frac{\| \mathbf{x} - \mathbf{\hat{x}} \|_2}{\| \mathbf{x} \|_2},
\]
where $\mathbf{x}$ denotes the original uncompressed data and $\mathbf{\hat{x}}$ denotes the reconstructed data.

\paragraph*{ROI Evaluation}
We conducted multiple experiments to assess the effectiveness of our approach.  Our network simultaneously predicts TC and AR. For each of these cases, we tried two UNet structures with eight layers and 13 layers, respectively.  We also varied the the input variables that were used for the prediction:

\begin{figure}
  \centering

  \begin{subfigure}[b]{0.24\textwidth}
    \includegraphics[width=\linewidth]{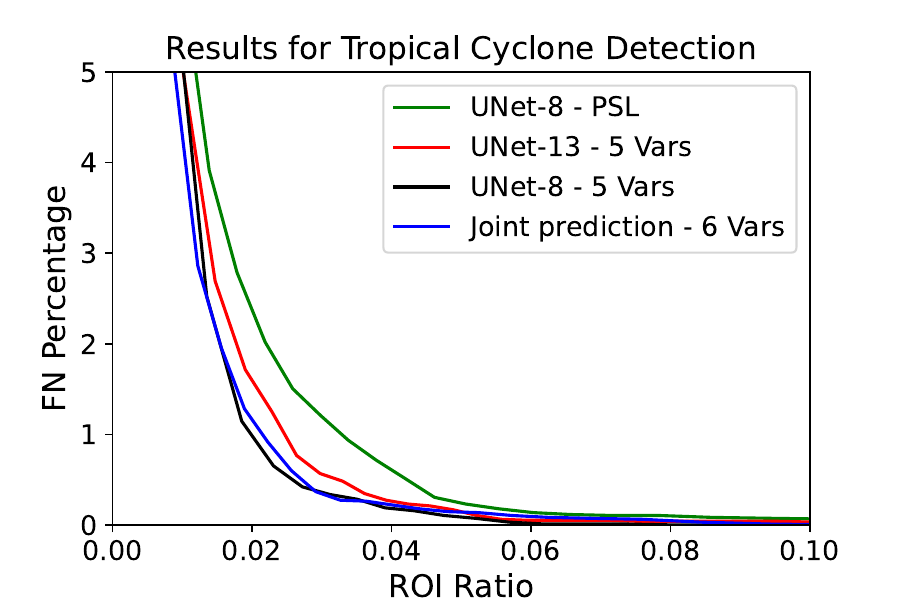}
    \label{fig:subfig1}
  \end{subfigure}
  \begin{subfigure}[b]{0.24\textwidth}
    \includegraphics[width=\linewidth]{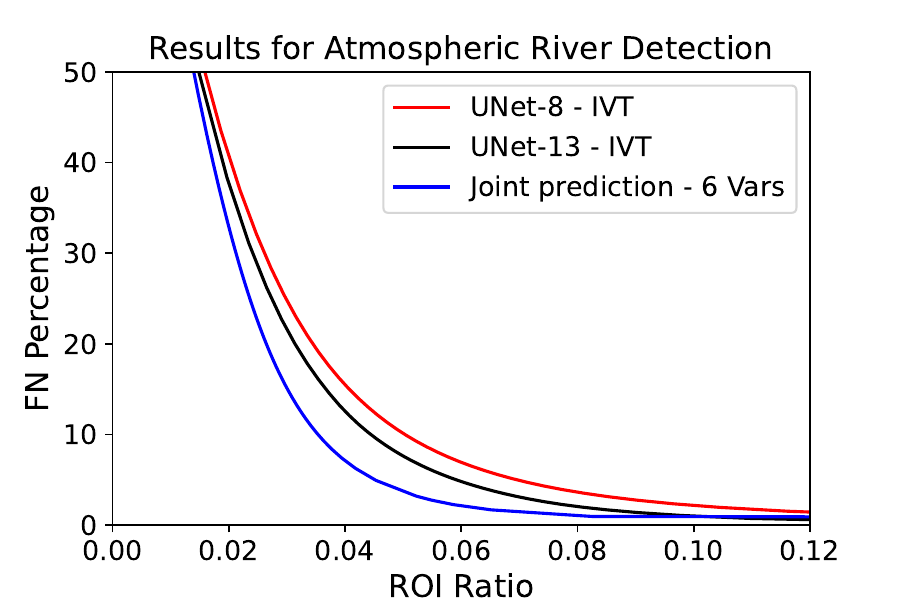}
    \label{fig:subfig2}
  \end{subfigure}
 \vspace*{-0.5cm}
  \caption{ROI ratio versus false negative ratio curve for TC and AR. These results show that a NN-based prediction approach for both TC and AR can achieve close to zero false negatives while keeping the ROIs to be less than 12 percent. }
  \label{fig:entire_figure}
   \vspace*{-0.5cm}
\end{figure}

\begin{compactitem}

    \item  `PSL' and `IVT' are the most important variables for TC detection and AR detection, respectively. For experiments marked with `PSL' or `IVT', we utilized only one variable as input.
    \item  `5 Vars' corresponds to the use of five variables: PSL, T500, T200, UBOT, and VBOT.
    \item All variables correspond to using all the six variables as input to jointly detect ARs and TCs.
\end{compactitem}

A comparison of ROI ratio and FN for different approaches is shown in Figure \ref{fig:entire_figure}.
 For TC detection, our method achieves a false negative ratio of 0, coupled with an ROI ratio of approximately 0.1. For AR detection, our approach attains an FN of less than $0.9\%$ for a corresponding ROI ratio of 0.08. In comparison to Qian et al. \cite{gong2022region}, which achieves a $0.7\%$ false negative ratio with a 0.22 ROI ratio for TC and a $5.8\%$ false negative ratio with a 0.36 ROI ratio for AR, our method exhibits substantial improvements in both TC and AR detection accuracy with lower ROI ratio.  Thus, by using our  approach, where the QoI are provided by existing application, we are able to more precisely determine the regions where they are present. At this juncture, any approach that uses differential compression should generally use higher compression in areas outside ROIs and should result in an overall better compression. We want to clarify again that the purpose of using the prediction model (that uses historical QoI) is only for determining ROIs. In the following sections, we show that both PD error and QoI errors are bounded. All the results are on unseen data used by the prediction mode.

\begin{table*}[htbp]

    \begin{minipage}{.66\linewidth}
    \begin{subtable}[b]{\linewidth}
    \scriptsize
        \centering
        \begin{tabular}{|c|c|c|c|c|c|c|c|c|}
        \hline
        \multirow{2}{*}{Method}& \multirow{2}{*}{ROI Ratio} & \multicolumn{6}{|c|}{Compression Ratio} &\multirow{2}{*}{TC Error Rate} \\
        \cline{3-8}
         & & PSL& T200 & T500 & VBOT & UBOT & Overall & \\
        \hline
        {Qian et al.\cite{gong2022region}}& 0.22 & 10.8 & 14.6& 17.1 & 18.3& 18.0 &15.2  & $0.38\% $\\
        
        \hline
        {MGARD-Uniform}&  - & 18.3& 21.6 & 22.6 & 26.1 & 24.7& 22.1 & $1.7 \%$\\

        \hline
        {MGARD-UNet}& 0.07  &  20.9& 48.7& 51.3& 51.5 & 50.1& 39.4 & $0.29 \%$\\

        \hline
        {GAE-Uniform}& - &20.7 & 22.4 & 20.3&  21.5& 20.3&  21.0& $1.2 \%$\\
        
        \hline
        {GAE-UNet}& 0.07  &54.3 & 63.0 &  66.8 & 66.3&  53.0&  60.1 & $0.29 \%$\\
        \hline
        \end{tabular}
        \caption{QoI preservation results for TC.}
        \label{tab:tc}
    \end{subtable}
    \end{minipage}
  \begin{minipage}{.3\linewidth}
    \begin{subtable}[b]{\linewidth}
        \centering
          \scriptsize
        \begin{tabular}{|c|c|c|c|c|}
        \hline
        
        {Method} & ROI Ratio & CR &  IoU \\  
        \hline
       

        {Qian et al.\cite{gong2022region}} & 0.37 & 9.1 & $99.3\%$\\

        \hline

        MGARD-Uniform & - & 12.1 & $99.1 \%$ \\
        \hline
        MGARD-UNet & 0.22 & 14.4 & $99.2 \%$ \\

        \hline
        
        GAE-Uniform & - & 17.4 & $99.2 \%$ \\

        \hline

        GAE-UNet & 0.22 & 23.4 & $99.4 \%$ \\ 

        \hline
        \end{tabular}
        
        \caption{QoI preservation results for AR.}
        \label{tab:ar}
    \end{subtable}
        \end{minipage}
    
    \caption{QoI preservation results for TC and AR. The `Uniform' corresponds to compressing the entire region with the same error bound. The `UNet' corresponds to differential compression where ROIs identified by UNet are compressed with lower error bound than other regions. These results show that our method has a significantly lower ROI and a higher compression ratio, but achieves lower QoI error as compared to previous work.}
    \label{tab:qoi}
     \vspace*{-0.5cm}
\end{table*}

\paragraph*{QoI Evaluation}

We applied GAE to perform differential compression and decompression on climate data with the ROIs identified by UNet model. Subsequently, we utilized TempestExtremes on both the reconstructed and original climate data, followed by a comparison of the results for TC and AR. TC preservation was assessed using the TC error rate, while AR preservation was evaluated using the Intersection over Union (IoU), as discussed in the previous section.

\begin{figure}[htbp]
  \centering
  \begin{subfigure}[b]{0.23\textwidth}
    \includegraphics[width=\textwidth]{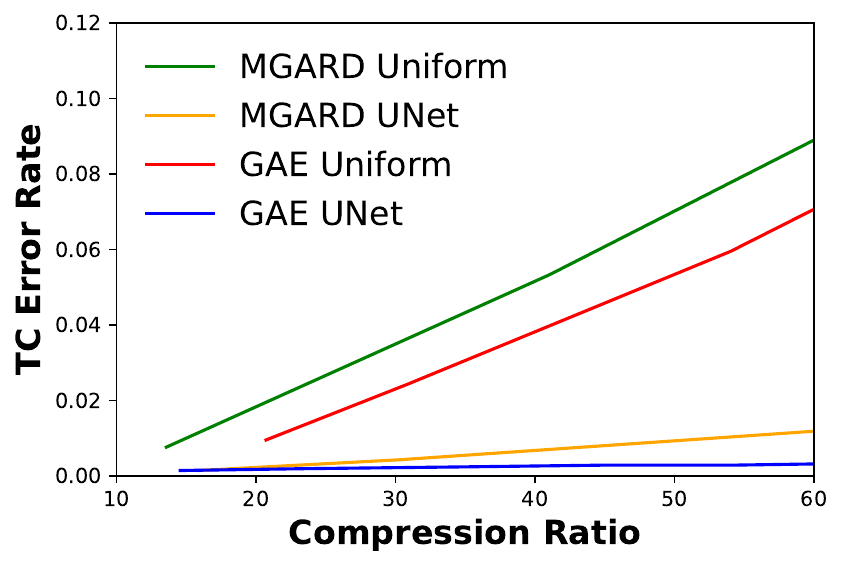}
    \caption{TC Error Rate on PSL data}
    \label{qoi:psl}
  \end{subfigure}
  \hfill
  \begin{subfigure}[b]{0.23\textwidth}
    \includegraphics[width=\textwidth]{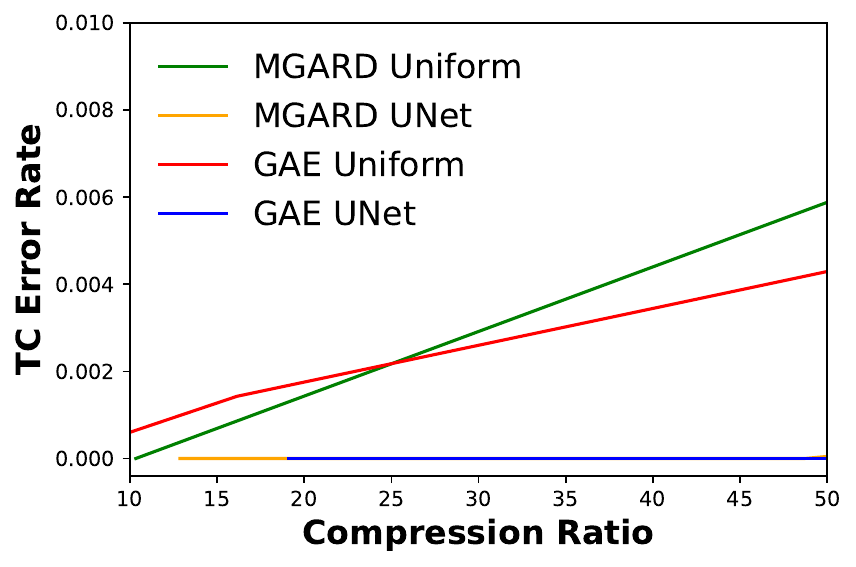}
    \caption{TC Error Rate on T200 data}
    \label{qoi:t200}
  \end{subfigure}
  
  \medskip
  \begin{subfigure}[b]{0.23\textwidth}
    \includegraphics[width=\textwidth]{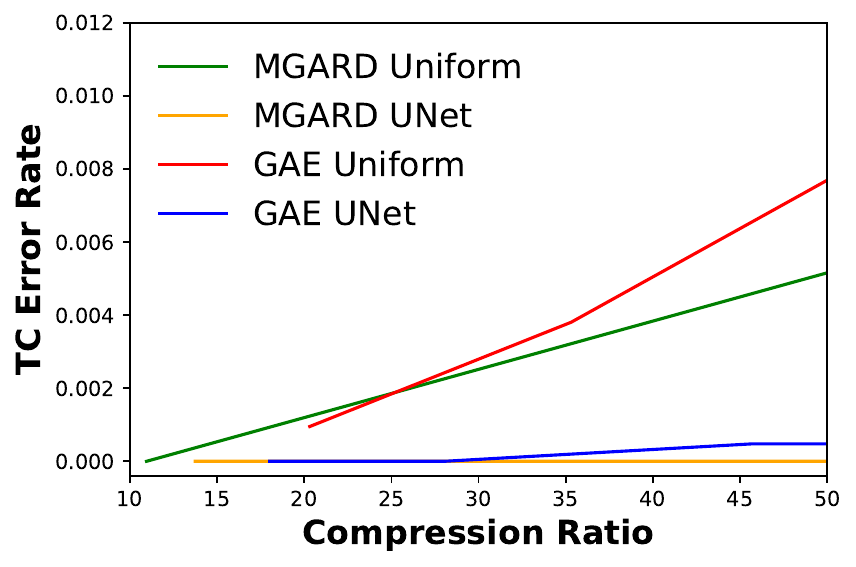}
    \caption{TC Error Rate on T500 data}
    \label{qoi:t500}
  \end{subfigure}
  \hfill
  \begin{subfigure}[b]{0.23\textwidth}
    \includegraphics[width=\textwidth]{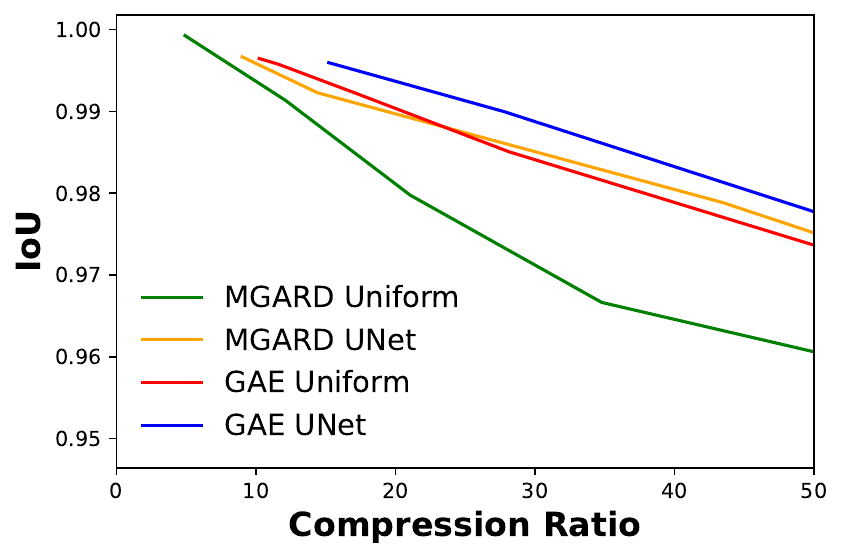}
    \caption{AR IoU on IVT data}
    \label{qoi:ar}
  \end{subfigure}
  \caption{QoI error for TC and AR. These results show that GAE with UNet is able to achieve higher compression ratio while achieving comparable or better accuracy to other methods.}
  \label{fig:tc_ar}
   \vspace*{-0.40cm}
\end{figure}

In our TC comparison, we conducted a series of experiments detailed in Table \ref{tab:tc}, comparing GAE with MGARD across various experimental setups. Experiments labeled `Uniform' employed a consistent error bound across all regions, while those marked `UNet' utilized region-adaptive error bounds. Qian et al. \cite{gong2022region} utilized region-adaptive compression using MGARD with ROIs detected by MGARD coefficients. When comparing MGARD-Uniform with GAE-Uniform at similar compression ratios, GAE achieved a lower TC error rate. Similarly, comparing MGARD-UNet with the approach by Qian et al. \cite{gong2022region}, UNet notably improved compression ratio by over 2 times while achieving a lower TC error rate. Moreover, in the comparison between MGARD-UNet and GAE-UNet, GAE further enhanced the compression ratio by approximately $50 \%$ while maintaining the same TC error rate.
Additionally, we present the compression ratio versus TC error rate curve, as depicted in Figure \ref{fig:tc_ar} (a-c). We specifically choose PSL, T200, and T500 as examples, given their significant influence on TC detection. Region-adaptive compression with UNet substantially outperformed uniform compression in terms of TC preservation. When comparing GAE with MGARD, GAE exhibited comparable or superior performance in both uniform and region-adaptive compression scenarios.

Similar trends are evident in the AR comparison, as summarized in Table \ref{tab:ar} and illustrated in Figure \ref{fig:tc_ar} (d). When comparing Qian et al.'s approach \cite{gong2022region} with MGARD-UNet at a close IoU, leveraging ROIs identified by UNet leads to a compression ratio improvement of over $50\%$. Further comparison between MGARD-UNet and GAE-UNet demonstrates an additional enhancement of approximately $60\%$ in compression ratio with the integration of GAE. By combining the advantages of GAE and UNet, GAE-UNet surpasses the performance of Qian et al. \cite{gong2022region} by 1.5 times in terms of compression ratio.


\begin{figure*}[htbp]
  \centering
  \begin{subfigure}[b]{0.24\textwidth}
    \includegraphics[width=\textwidth]{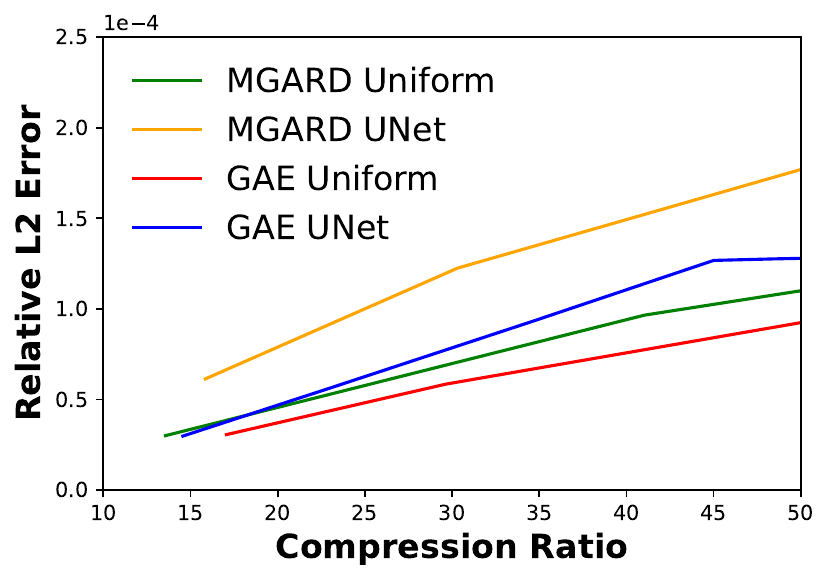}
    \caption{PD error on PSL}
    \label{pd:psl}
  \end{subfigure}
  \hfill
  \begin{subfigure}[b]{0.24\textwidth}
    \includegraphics[width=\textwidth]{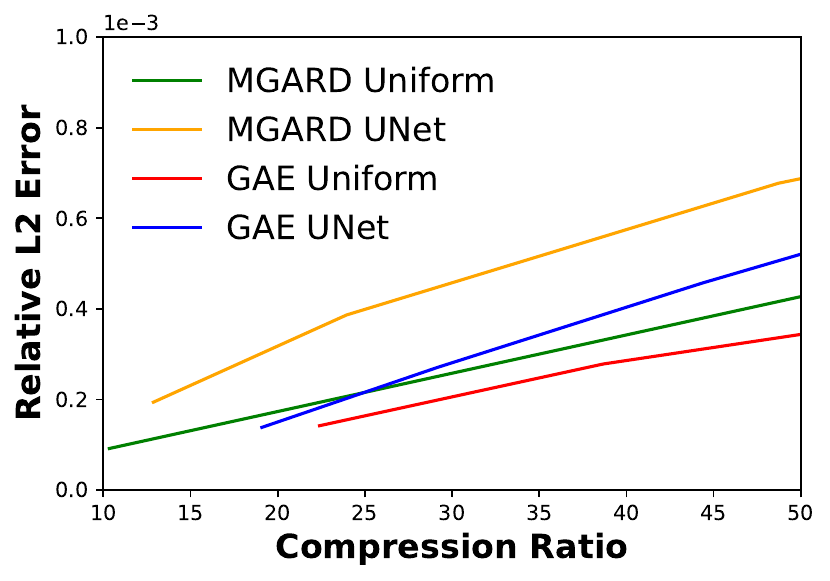}
    \caption{PD error on T200}
    \label{pd:t200}
  \end{subfigure}
  \hfill
    \eatme{
  \begin{subfigure}[b]{0.23\textwidth}
  
    \includegraphics[width=\textwidth]{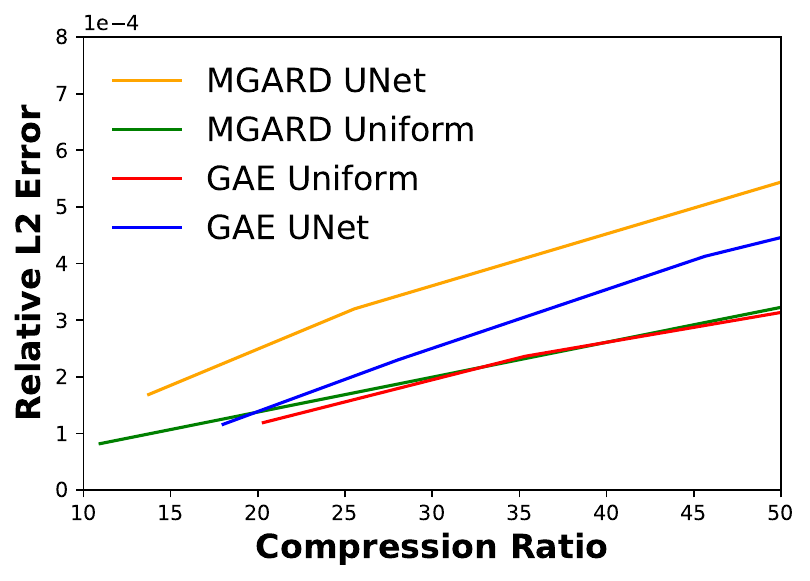}
    \caption{PD error on T500}
    \label{pd:t500}
  \end{subfigure}
  \hfill
  \begin{subfigure}[b]{0.24\textwidth}
    \includegraphics[width=\textwidth]{results_pd/VBOT_PD_nt50.pdf}
    \caption{PD error on VBOT}
    \label{pd:vbot}
  \end{subfigure}
    }
  \hfill
  \begin{subfigure}[b]{0.24\textwidth}
  
    \includegraphics[width=\textwidth]{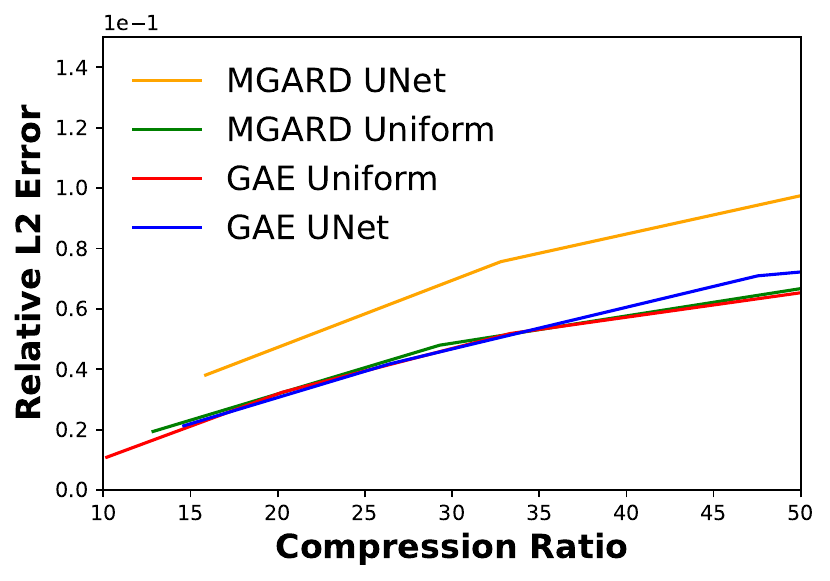}
    \caption{PD error on UBOT}
    \label{pd:ubot}
  \end{subfigure}
  \hfill
  \begin{subfigure}[b]{0.24\textwidth}
    \includegraphics[width=\textwidth]{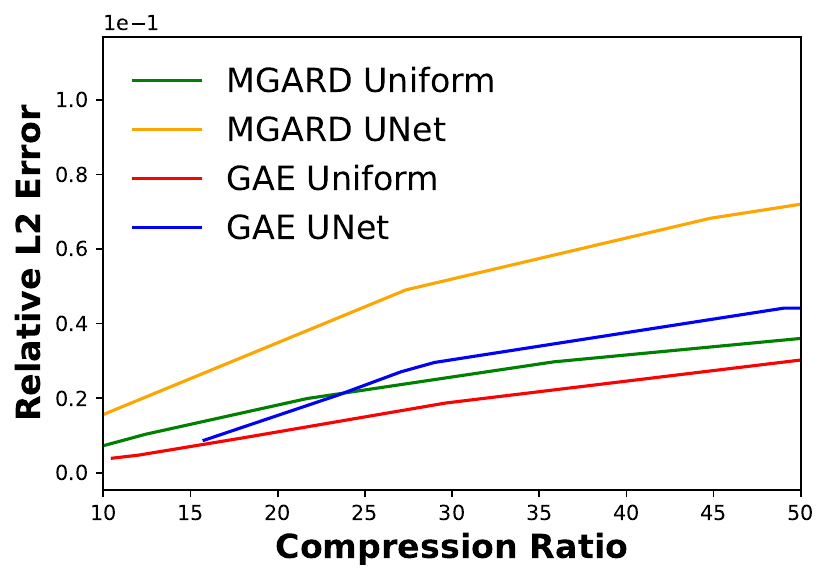}
    \caption{PD error on TVQ}
    \label{pd:tvq}
  \end{subfigure}
  \hfill
    \eatme{
  \begin{subfigure}[b]{0.24\textwidth}
 
    \includegraphics[width=\textwidth]{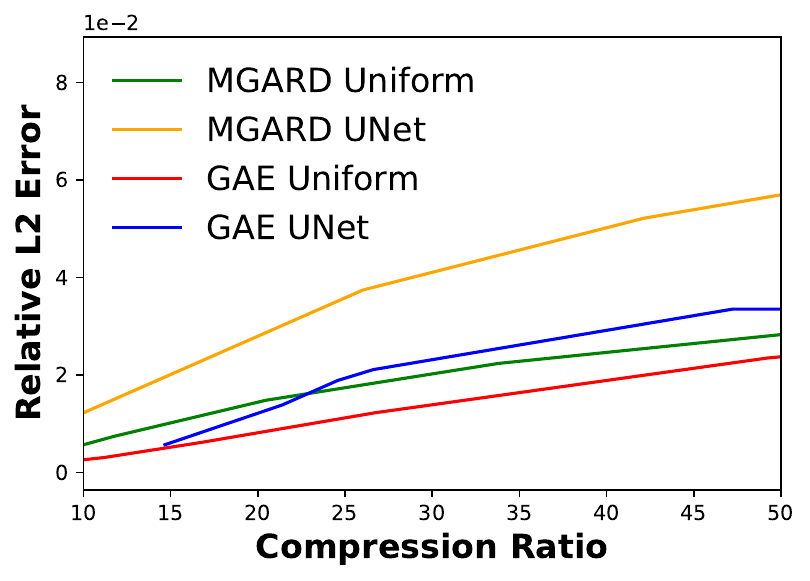}
    \caption{PD error on TUQ}
    \label{pd:tuq}
  \end{subfigure}
}

  \caption{Comparison of PD error for four variables using the four methods described in the paper. The other three variables have similar relative behavior for the four methods. Overall, GAE-based approaches provide higher compression than other methods for the same error levels and comparable QoI. The QoI are for a subset of these variables, presented in Figure \ref{fig:tc_ar}.}
  \label{fig:pd_error}
   \vspace*{-0.5cm}
  
\end{figure*}

\paragraph*{PD Error} To evaluate the performance of GAE on primary data, we conducted several comparisons between different experimental setups. The comparisons were conducted on all seven climate variables, but we only show four here due to the page constraints. We compared MGARD with GAE and also compared uniform compression with region-adaptive compression. The experimental results are shown in Figure \ref{fig:pd_error}.
When measuring the error globally, uniform compression achieves lower PD error than region-adaptive compression. This finding holds true for both MGARD and GAE. When comparing MGARD with GAE, it's notable that GAE consistently achieves a similar or lower PD error than MGARD at the same compression ratio, irrespective of whether uniform or region-adaptive compression is employed.

\section{Related Work}\label{relwork}
Error-bounded lossy compression has drawn increasing attention as it allows more significant data reduction and ensures the quality of the reconstructed data to meet the user's request. 
In general, lossy compressors reduce data through multiple steps including de-correlation, quantization/thresholding, and lossless compression. 
Based on the de-correlation algorithms employed by lossy compressors, they can be categorized into prediction-based \cite{di2016fast,liang2018error,tian2020cusz}, transformation-based \cite{diffenderfer2019error,fox2020stability,lindstrom2014fixed}, and decomposition-based methods \cite{ainsworth2019multilevel,liang2021mgard+,ainsworth2019qoi,chen2021scalable}. 
Most lossy compressors \cite{di2016fast, gong2022region, lindstrom2014fixed} only support applying uniform error bounds (e.g., $L^2$ or $L^\infty$) across the entire data space. Previous research on varying the compression quality for dynamically changing data space typically shows a need to include region segmentation types of methods \cite{rahman2022dynamic,uddehal2023image,akutsu2020end} and compressing foreground and background data separately. For ROIs with irregular shapes, partitioning them from background can result in oversized bounding boxes or fragmented data pieces, which leads to suboptimal compression ratios. Recently, Liang et al. \cite{jiao2022toward,liang2022toward} derived local error bounds which can be used to preserve critical point features for a prediction-based compressor, SZ. However, their method demands a mask of ROIs and different reconstruction algorithms to assemble pixels compressed with varied error bounds. 
In previous work~\cite{gong2022region,gong2023spatiotemporally}, Gong et al. employ the multilevel coefficients from MGARD decomposition to assess local data turbulence. QoI are treated as variational features, and detected using a hierarchical mesh refinement approach which tracks the clusters of large-magnitude coefficients. The authors only categorize the QoI into nodal-based and areal-based features when making different mesh refinement parameters choices. 

\section{Conclusions}\label{conclusion}

Scientists conduct large-scale simulations to compute derived quantities from primary data, and it is crucial for them to find compression techniques that maintain low or bounded errors on these derived quantities called quantities of interest (QoI). Additionally, for many spatiotemporal applications these QoI are binary in nature (presence or absence of a physical phenomenon). 
We demonstrated a multilevel compression approach that first uses neural-network-based techniques to derive regions where QoI are highly likely to be present. Then, an autoencoder with error bound guarantee is proposed to perform differential compression based on these regions.
This results in overall high compression ratios while still achieving downstream goals of simulation or data collections.

Experimental results are presented for climate data generated from the E3SM simulation model for downstream QoI such as tropical cyclone and atmospheric river detection and tracking. 
Compared to previous research\cite{gong2022region}, our method achieves a significantly higher compression ratio while simultaneously achieving a lower false negative (FN) ratio for both TC and AR detection.





\bibliographystyle{IEEEtran}
\bibliography{ref.bib}

\end{document}